\documentclass[conference]{IEEEtran}
\IEEEoverridecommandlockouts

\usepackage{cite}
\usepackage{amsmath,amssymb,amsfonts}
\usepackage{algorithm}
\usepackage{algpseudocode}
\usepackage{booktabs}
\usepackage{multirow}
\usepackage{subcaption}
\usepackage{graphicx}
\usepackage{textcomp}
\usepackage{xcolor}

\def\BibTeX{{\rm B\kern-.05em{\sc i\kern-.025em b}\kern-.08em
    T\kern-.1667em\lower.7ex\hbox{E}\kern-.125emX}}

\algdef{SE}[DOWHILE]{Do}{doWhile}{\algorithmicdo}[1]{\algorithmicwhile\ #1}

\makeatletter
\newcommand{\linebreakand}{%
  \end{@IEEEauthorhalign}%
  \hfill\mbox{}\par
  \mbox{}\hfill\begin{@IEEEauthorhalign}%
}
\makeatother

\begin{document}

\title{Physics-Aware Heterogeneous GNN Architecture for Real-Time BESS Optimization in Unbalanced Distribution Systems
\thanks{Corresponding author: Jun Cao, Email: jun.cao@list.lu. This work was partially funded by FNR CORE project LEAP (17042283), FNR CORE C24/IS/18942843, and the European Commission under projects iSTENTORE (HORIZON-CL5-2022-D3-01, ID-101096787) and We-Forming (HORIZON-CL5-2022-D4-02-04, ID-101123556).}%
}

\author{%
\IEEEauthorblockN{Aoxiang MA}
\IEEEauthorblockA{\textit{LIST},~\IEEEmembership{Student Member,~IEEE}\\
Esch-Belval, Luxembourg\\
https://orcid.org/0009-0005-9553-5650}
\and
\IEEEauthorblockN{Salah GHAMIZI}
\IEEEauthorblockA{\textit{Luxembourg Institute of Health}\\
\textit{University of Luxembourg/SnT}, Luxembourg\\
https://orcid.org/0000-0002-0738-8250}
\and
\IEEEauthorblockN{Jun CAO}
\IEEEauthorblockA{\textit{LIST},~\IEEEmembership{Member,~IEEE}\\
Esch-Belval, Luxembourg\\
https://orcid.org/0000-0001-5099-9914}
\linebreakand
\IEEEauthorblockN{Pedro RODRIGUEZ CORTES}
\IEEEauthorblockA{\textit{LIST, University of Luxembourg},~\IEEEmembership{Fellow,~IEEE}\\
Esch-Belval, Luxembourg\\
https://orcid.org/0000-0002-1865-0461}
}

\maketitle

\begin{abstract}
Battery energy storage systems (BESS) have become increasingly vital in three-phase unbalanced distribution grids for maintaining voltage stability and enabling optimal dispatch. However, existing deep learning approaches often lack explicit three-phase representation, making it difficult to accurately model phase-specific dynamics and enforce operational constraints—leading to infeasible dispatch solutions. This paper demonstrates that by embedding detailed three-phase grid information—including phase voltages, unbalanced loads, and BESS states—into heterogeneous graph nodes, diverse GNN architectures (GCN, GAT, GraphSAGE, GPS) can jointly predict network state variables with high accuracy. Moreover, a physics-informed loss function incorporates critical battery constraints—SoC and C-rate limits—via soft penalties during training. Experimental validation on the CIGRE 18-bus distribution system shows that this embedding-loss approach achieves low prediction errors, with bus voltage MSEs of 6.92e-07 (GCN), 1.21e-06 (GAT), 3.29e-05 (GPS), and 9.04e-07 (SAGE). Importantly, the physics-informed method ensures nearly zero SoC and C-rate constraint violations, confirming its effectiveness for reliable, constraint-compliant dispatch.

\end{abstract}


\section{Introduction}
Three-phase unbalanced distribution grids present unique challenges for real-time battery optimal operational dispatch. Unlike transmission systems characterized by balanced three-phase conditions, distribution grids exhibit significant phase imbalances due to uneven load distribution across phases, creating substantial phase load differences~\cite{sun2023analysis}. These imbalances create significant phase-to-phase coupling effects via mutual impedances that single-phase approximations fail to capture~\cite{kersting2010distribution}. The integration of distributed energy resources and battery energy storage systems further complicates the operation, requiring both accurate multi-variable prediction and constraint-aware dispatch strategies~\cite{gangwar2024energy}.

Traditional approaches rely on three-phase OPF solvers combined with a Multi-step optimizer for battery optimal dispatch~\cite{opends}. Although these methods are rigorous and physically sound, they require long computation time per optimization cycle, fundamentally limiting real-time applications in dynamic grid environments. Graph neural networks (GNNs) have emerged as scalable alternatives, exploiting grid topology through localized parameterization where parameters are defined on nodes and edges rather than globally~\cite{owerko2020optimal}. This localized approach enables superior performance for predicting voltage magnitudes, phase angles, and power injections compared to fully connected architectures~\cite{donon2020neural}. Recent advances have extended GNNs to multi-task frameworks, simultaneously forecasting multiple power system variables while capturing spatial dependencies in complex grid topologies~\cite{hansen2021power}.

Heterogeneous GNNs have advanced constraint-aware control capabilities. Ghamizi et al.~\cite{ghamizi2024opf} proposed OPF-HGNN with type-specific embeddings and differentiable constraints, achieving 98\% constraint satisfaction on balanced networks with two orders of magnitude speedup. Ma et al.~\cite{ma2025hil} validated real-time operation through hardware-in-the-loop (HIL) simulations using the SafePowerGraph-HIL framework, demonstrating practical feasibility for grid deployment.

However, existing HIL-tested heterogeneous GNN approaches remain limited to single-phase or balanced-condition scenarios~\cite{ma2025hil}, lacking explicit battery constraint enforcement in unbalanced systems where SoC and C-rate violations present critical operational risks. While PowerFlowMultiNet~\cite{ghamizi2024multigraph} addresses unbalanced three-phase power flow prediction using multigraph representations, and physics-informed approaches~\cite{qin2025physics} enable dynamic voltage regulation, these works focus primarily on power flow prediction and voltage control rather than constraint-aware battery dispatch in unbalanced distribution grids.

This paper presents the first physics-informed heterogeneous GNN framework specifically designed for constraint-aware battery dispatch in three-phase unbalanced grids. The main contributions are:

\begin{enumerate}
\item \textbf{Explicit three-phase grid embedding}: All grid components are uniformly encoded with three-phase information, enabling GNN layers to learn phase-specific dynamics and phase-coupled interactions without decomposition.

\item \textbf{Physics-informed loss for constraint satisfaction}: A loss function that combines prediction accuracy with SoC and C-rate constraint enforcement, enabling the model to learn constraint-aware dispatch that guarantees nearly zero violations at deployment.
\end{enumerate}

The remainder of this paper is organized as follows: Section II presents the physics-informed heterogeneous GNN framework, including three-phase grid embedding and multi-task loss formulation. Section III validates the approach on the CIGRE 18-node testing system. Section IV concludes and outlines future work.

\section{Physics-Aware Heterogeneous GNN Architecture}
\subsection{Problem Formulation}



This work addresses multi-task optimal constraint-aware battery dispatch in three-phase unbalanced distribution grids through a physics-informed heterogeneous GNN framework, jointly predicting bus voltages (magnitude and phase across phases A, B, C), external grid power injection (P, Q), and optimal battery charge/discharge power while satisfying operational constraints.
The control objective maximizes net revenue:
$$\max \sum_{t=1}^{T} \lambda_t P_t^{\text{grid}} \Delta t$$
where $\lambda_t$ is the electricity price and $P_t^{\text{grid}}$ is the grid power exchange.

The three-phase AC power flow equations govern grid dynamics:
$$p_i^\phi + q_i^\phi j = v_i^\phi (Xv)_i^{\phi *}, \quad \phi \in \{a,b,c\}$$
where $p_i^\phi, q_i^\phi$ are active and reactive power injections at bus $i$ phase $\phi$, $v_i^\phi$ is the complex voltage, $X$ is the three-phase admittance matrix (X-bus), and the subscript $*$ denotes complex conjugate. Power balance at each bus incorporates load, battery, and PV contributions:
$$p_i^\phi = p_i^{\text{load}} + p_i^{\text{bess}} + p_i^{\text{pv}}$$

Battery state-of-charge dynamics follow:
$$\text{SOC}_{t+1} = \text{SOC}_t + \frac{\eta_c p_t^{\text{ch}} - p_t^{\text{dis}}/\eta_d}{E_{\text{rated}}} \Delta t$$
where $p_t^{\text{ch}}, p_t^{\text{dis}}$ are charging and discharging power, $\eta_c, \eta_d$ are charging and discharging efficiencies, and $E_{\text{rated}}$ is the rated battery capacity. Operational constraints enforce SOC and voltage limits:
$$\text{SOC}_{\min} \leq \text{SOC}_t \leq \text{SOC}_{\max}, \quad V_{\min} \leq |v_i^\phi| \leq V_{\max}$$

Battery power is limited by both rated power and C-rate (maximum charge/discharge rate as a multiple of capacity):
$$0 \leq p_t^{\text{ch}}, p_t^{\text{dis}} \leq \min\{P_{\text{rated}}, C_{\text{rate}} \cdot E_{\text{rated}}\}$$
with the implicit constraint that charging and discharging do not occur simultaneously.
\subsection{Unbalanced Grid Graph Construction}

\subsubsection{Graph Embedding}
\begin{figure}[htbp]
  \centering
  \includegraphics[width=\linewidth]{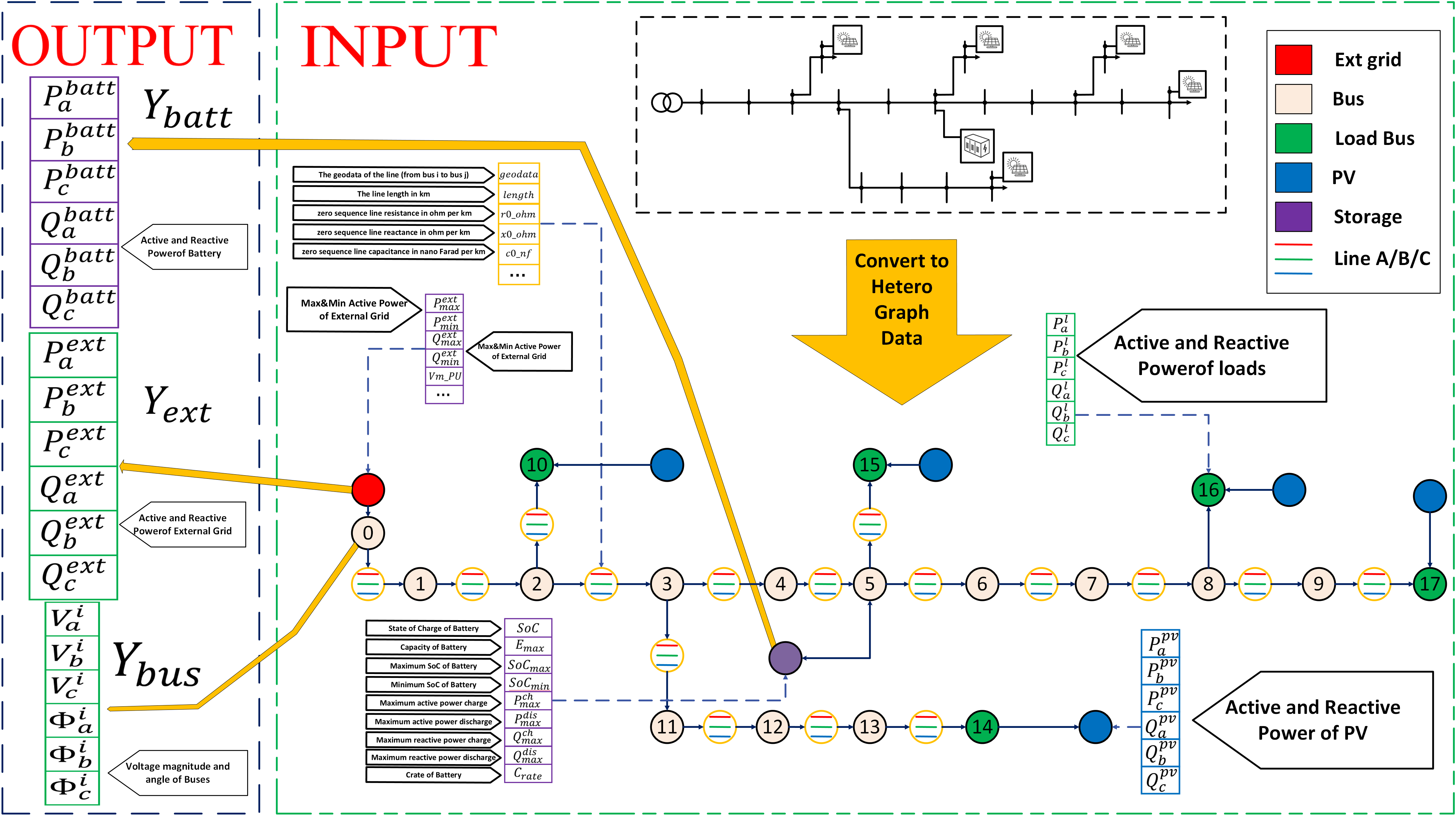}
  \caption{Graph embedding for CIGRE 18-bus grid}
  \label{fig:pes26}
  \vspace{-1em}
\end{figure}

A three-phase distribution grid is modeled as a heterogeneous graph $G = (\mathcal{N}_T, \mathcal{E})$ where $\mathcal{N}_\tau$ is the set of nodes for each node type $\tau \in \{\text{bus, load, line, ...}\}$, $\mathcal{E}$ represents the set of grid connections.
The graph encodes the grid topology and static parameters as input node features, while voltage magnitudes, angles, active power and reactive power are predicted as regression targets, as shown in Fig. \ref{fig:pes26}.

\textbf{Bus nodes} encode static parameters: $$\mathbf{x}_{\text{bus}} = [V_{\text{rated}}, V_{\max}, V_{\min}, \text{type}]$$ where $V_{\text{rated}}$ is rated voltage, $V_{\min}$ and $V_{\max}$ are bounds, and type indicates bus type (PQ, PV, or slack).

\textbf{Storage nodes} encode the battery state, capacity, and operating limits as:
$$
\mathbf{x}_{\text{storage}} =
\begin{bmatrix}
\text{SoC},~ E_{\max},~ \text{SoC}_{\max},~ \text{SoC}_{\min},~ P_{\max}^{ch}, \\
P_{\max}^{dis},~ Q_{\max}^{ch},~ Q_{\max}^{dis},~ C_{\text{rate}}
\end{bmatrix}
$$

where $\text{SoC}$ denotes the current state of charge, 
$E_{\max}$ is the nominal capacity, 
$\text{SoC}_{\max}$ and $\text{SoC}_{\min}$ are charge limits, 
$P_{\max}^{ch}$ and $P_{\max}^{dis}$ represent maximum active power for charge and discharge, 
$Q_{\max}^{ch}$ and $Q_{\max}^{dis}$ are corresponding reactive power limits, 
and $C_{\text{rate}}$ denotes the charge/discharge rate constraint.

\textbf{External grid nodes} encode grid connection limits:
$$
\mathbf{x}_{ext\_grid} = [P_{\min}^{ext}, P_{\max}^{ext}, Q_{\min}^{ext}, Q_{\max}^{ext}]
$$
\textbf{Load nodes} encode three-phase power demands $(P_a^{load}, Q_a^{load}, P_b^{load}, Q_b^{load}, P_c^{load}, Q_c^{load})$ as input features representing the current load profile.



\textbf{Line nodes} encode three-phase transmission line parameters including series impedance and shunt admittance. The feature representation captures both phase-symmetric properties and phase-coupling effects:

$$
\mathbf{x}_{ij}^{\text{line}} =
\begin{bmatrix}
r_{\text{ohm}}, x_{\text{ohm}}, g_{\text{us}}, b_{\text{us}}, c_{\text{par}}, df, \\
R_a, X_a, R_b, X_b, R_c, X_c, G_{ab}, G_{bc}, G_{ca}
\end{bmatrix}
$$
where the first six elements are line-level properties (resistance and reactance per-km, ground conductance and susceptance per-km, parallel circuits, dielectric loss factor) that apply identically across all three phases. The remaining nine elements explicitly encode three-phase-specific impedances: series resistance and reactance for each phase ($R_{\phi}, X_{\phi}$ for $\phi \in \{a,b,c\}$), and mutual admittances between phases ($G_{ab}, G_{bc}, G_{ca}$) that quantify phase-coupling effects.

We model lines as nodes, given the effectiveness and superiority of this representation~\cite{ghamizi2024multigraph}. This heterogeneous approach enables the GNN to learn how line impedances and mutual coupling terms affect voltage propagation across phases.

\subsubsection{Prediction Targets}

The framework jointly optimizes three regression objectives:

\textbf{Bus voltage} ($\mathbf{y}_{bus}$): Three-phase voltage magnitudes and angles
$$
\mathbf{y}_{bus} = [V_a^{mag}, V_a^{ang}, V_b^{mag}, V_b^{ang}, V_c^{mag}, V_c^{ang}] \in \mathbb{R}^6
$$
\textbf{External grid power} ($\mathbf{y}_{ext}$): Three-phase active and reactive power at grid connection points
$$
\mathbf{y}_{ext} = [P_a^{ext}, Q_a^{ext}, P_b^{ext}, Q_b^{ext}, P_c^{ext}, Q_c^{ext}] \in \mathbb{R}^6
$$
\textbf{Battery dispatch} ($\mathbf{y}_{storage}$): Three-phase active and reactive power for storage units
$$
\mathbf{y}_{storage} = [P_a^{batt}, Q_a^{batt}, P_b^{batt}, Q_b^{batt}, P_c^{batt}, Q_c^{batt}] \in \mathbb{R}^6
$$

These predictions are learned through end-to-end training, where the model jointly optimizes bus voltages and angles, external grid active and reactive power, and battery dispatch with physics-informed constraints ensuring feasibility.
\subsection{Heterogeneous Graph Neural Network Architecture}

\subsubsection{Type-Specific Message Passing}

The heterogeneous GNN employs type-specific weight matrices for message passing. The aggregated message to node $i$ at layer $\ell$ combines neighbor features and edge information:
$$\mathbf{m}_{i}^{(\ell)} = \sum_{j \in \mathcal{N}(i)} W_{\tau_j \to \tau_i}^{(\ell)} \mathbf{x}_j^{(\ell-1)} + \text{AGG}(\mathbf{e}_{ij})$$

where $W_{\tau_j \to \tau_i}^{(\ell)}$ is a type-specific weight matrix transforming features from node type $\tau_j$ to $\tau_i$, and $\text{AGG}(\mathbf{e}_{ij})$ aggregates edge features via summation, mean, or attention-weighted combination.

Node embeddings integrate self and neighbor information:
$$\mathbf{x}_i^{(\ell)} = \sigma\left(W_{\tau_i}^{\text{self},(\ell)} \mathbf{x}_i^{(\ell-1)} + W_{\tau_i}^{\text{agg},(\ell)} \mathbf{m}_{i}^{(\ell)}\right)$$

where $W_{\tau_i}^{\text{self},(\ell)}$ processes node's own features, $W_{\tau_i}^{\text{agg},(\ell)}$ processes aggregated neighbor information, and $\sigma$ is ReLU activation. This layer-wise update repeats $K$ times, progressively building embeddings $\mathbf{x}_i^{(K)}$ that integrate multi-hop neighborhood information.
Heterogeneous GNNs are essential in power system applications because system components have inherently different properties and serve distinct operational purposes. Type-specific weight matrices $W_{\tau_j \to \tau_i}^{(\ell)}$ enable specialized information routing between component types, whereas homogeneous GNNs fail to capture these functional distinctions.

\subsubsection{Multi-Task Output Prediction}

The framework employs three specialized output heads, each predicting a specific objective from the learned node embeddings:

$$\begin{pmatrix}
\hat{\mathbf{V}}_{\text{bus}} \\
\hat{\mathbf{P}}_{\text{ext}} \\
\hat{\mathbf{P}}_{\text{batt}}
\end{pmatrix} = \begin{pmatrix}
\text{head}_{\text{bus}}(\mathbf{x}_{bus}^{(K)}) \\
\text{head}_{\text{ext}}(\mathbf{x}_{ext}^{(K)}) \\
\text{head}_{\text{storage}}(\mathbf{x}_{storage}^{(K)})
\end{pmatrix}$$

The bus head predicts three-phase voltage magnitudes and angles, the external grid head predicts three-phase active power and reactive power, and the storage head predicts battery dispatch.


\subsection{Loss function with Physics-Aware Loss}

The total loss combines prediction accuracy with constraint satisfaction through a weighted multi-objective formulation:
$$
\mathcal{L}(\Theta) := \sum_{g \in \mathcal{G}} \left[ \left\| \mathbf{y}_g - \hat{\mathbf{y}}_g \right\|_2^2 + \lambda_{\text{phys}} \sum_{i \in \Omega_g} \text{penalty}(\mathbf{x}_g, \omega_i) \right]
\label{eq:total_loss}
$$
where $\mathcal{G}$ denotes the set of graphs (time steps), $\mathbf{y}_g$ is the ground truth multi-task target (bus V/A, external grid P/Q, battery P/Q), $\hat{\mathbf{y}}_g$ is the predicted output, and $\Omega_g$ represents the set of operational constraints for graph $g$. The hyperparameter $\lambda_{\text{phys}}$ weights the contribution of physics-informed penalty terms.


\subsubsection{Constraint Formulation}

The physics-informed penalty enforces operational constraints across three component types:

\textbf{Battery constraints:} Battery state-of-charge (SoC) evolves according to energy balance: $\text{SoC}_{t} = \text{SoC}_{t-1} - P_t^{\text{pred}}/E_{\max}$ with operational bounds $\text{SoC}_{\min} \leq \text{SoC}_{t} \leq \text{SoC}_{\max}$. Additionally, charge/discharge power must respect the C-rate limit: $|P_t^{\text{pred}}| \leq C_{\text{rate}} \cdot E_{\max}$.

\textbf{Bus voltage constraints:} Each bus must maintain voltage within acceptable ranges across all three phases: $V_{\min} \leq V_{\phi,t}^{\text{pred}} \leq V_{\max}$ for $\phi \in \{a,b,c\}$.

\textbf{External grid constraints:} Grid connection power must respect capacity limits: $P_{\min}^{\text{ext}} \leq P_{\phi,t}^{\text{ext,pred}} \leq P_{\max}^{\text{ext}}$ and $Q_{\min}^{\text{ext}} \leq Q_{\phi,t}^{\text{ext,pred}} \leq Q_{\max}^{\text{ext}}$ for each phase $\phi$.

\subsubsection{Penalty Function}

Rather than enforcing hard constraints during training (which complicates gradient flow), soft penalty terms guide the model toward constraint-aware predictions. The physics-informed penalty combines violation terms across all component types:

$$\text{penalty} = \mathcal{L}_{\text{SoC}} + \sum_{\phi \in \{a,b,c\}} \mathcal{L}_{V_{\phi}} + \sum_{\phi \in \{a,b,c\}} \left(\mathcal{L}_{P_{\phi}}^{\text{ext}} + \mathcal{L}_{Q_{\phi}}^{\text{ext}}\right)$$

where each loss term is defined as:

$$\mathcal{L}_{\text{SoC}} = \mathbb{E}\left[\left(\max\left(0, |\text{SoC} - [SoC_{\min}, SoC_{\max}]|\right)\right)^2\right]$$
$$\mathcal{L}_{V_{\phi}} = \mathbb{E}\left[\left(\max\left(0, |V_{\phi} - [V_{\min}, V_{\max}]|\right)\right)^2\right]$$
$$\mathcal{L}_{P_{\phi}}^{\text{ext}} = \mathbb{E}\left[\left(\max\left(0, |P_{\phi}^{\text{ext}} - [P_{\min}, P_{\max}]|\right)\right)^2\right]$$
$$\mathcal{L}_{Q_{\phi}}^{\text{ext}} = \mathbb{E}\left[\left(\max\left(0, |Q_{\phi}^{\text{ext}} - [Q_{\min}, Q_{\max}]|\right)\right)^2\right]$$

The physics-informed penalty aggregates the mean-squared violations of operational limits on the battery state of charge (SoC), per-phase voltages, and external grid powers. Each loss term measures how far a variable exceeds its allowable range, ensuring that the model learns to produce predictions consistent with physical and operational constraints. This formulation encourages constraint-aware behavior during training while maintaining flexibility for the learning process.

\section{Empirical Study}
\paragraph{Use case}
We evaluate our algorithm on 3-phase grid topologies: CIGRE 18 bus distribution grid with 1 modified battery, 5 unbalanced loads, and 5 unbalanced PVs.

\paragraph{Simulation tool} We use PandaPower\cite{thurner_pandapower_2018} for topology validation, 3-phase optimal power flow (OPF) ground truth generation. The performance study uses 8,000 training datasets and 2,000 validation datasets.

\begin{figure}[htbp]
  \centering
  \includegraphics[width=0.5\textwidth]{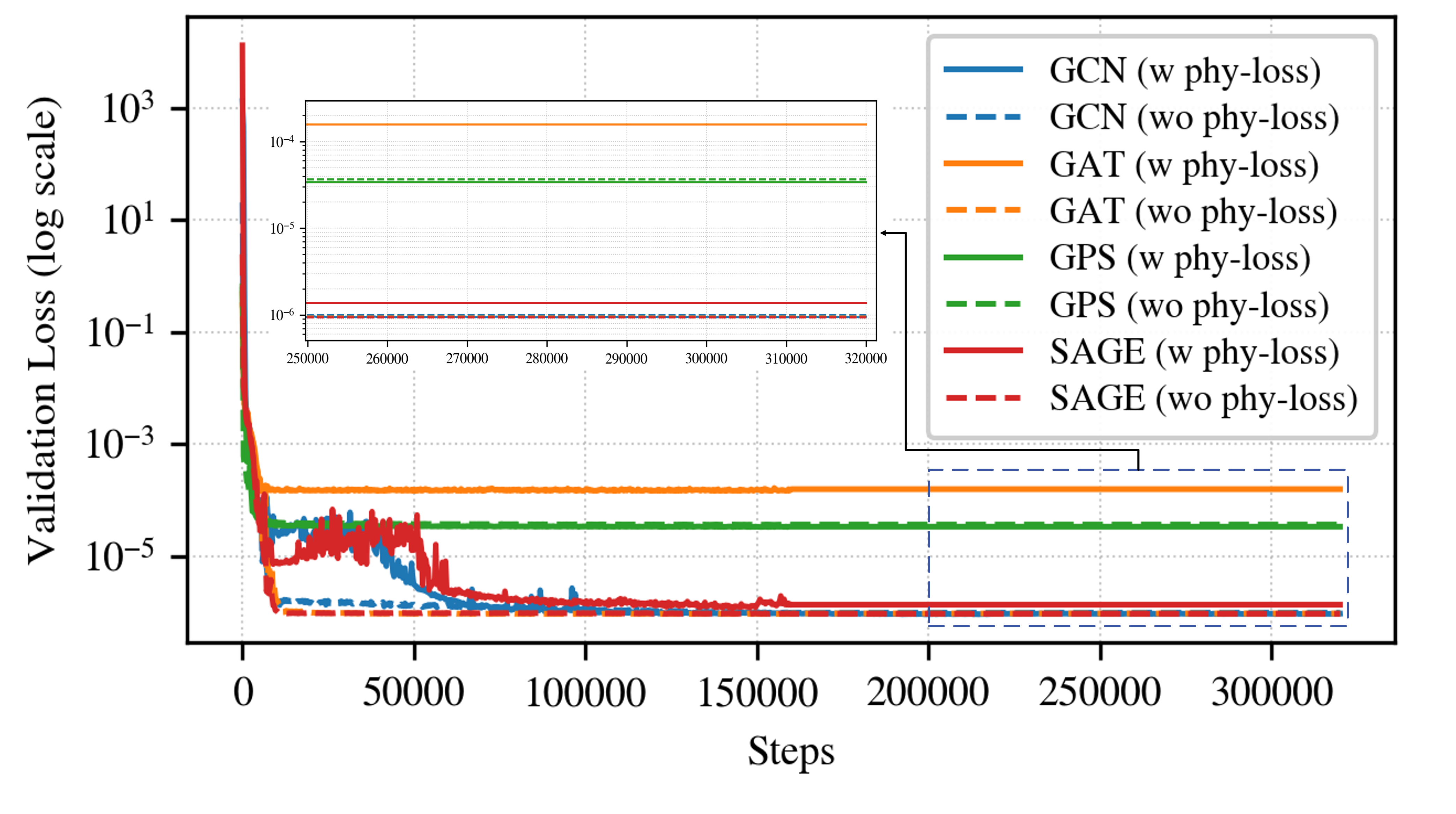}
  \caption{Validation losses for multi-task objectives across GCN, GAT, GPS, and SAGE architectures with and without physics-informed loss components}
  \label{fig:val_loss}
  \vspace{-1em}
\end{figure}

\paragraph{Multi-Task Validation Loss Convergence}
The framework jointly optimizes three regression objectives via multi-task learning. Figure~\ref{fig:val_loss} shows the validation loss convergence of the three prediction tasks—bus voltages, external grid power flows, and battery state-of-charge—over 2000 training epochs, comparing models trained with and without physics-informed loss terms.

We evaluate four HGNN architectures on three-phase unbalanced distribution grids: GCN, GAT, GraphSAGE, and—for the first time—GPS. All models converge reliably, confirming the effectiveness of explicit three-phase encoding. GPS yields slightly higher validation loss than GCN, GAT, and SAGE (e.g., GPS voltage loss: $3.29\times10^{-5}$ vs. GCN: $6.92\times10^{-7}$), but remains a promising option due to its scalability for larger grids.

\paragraph{Multi-Task Convergence Analysis}
Table~\ref{tab:boundary_loss_all_final} presents validation loss (MSE) across all three prediction tasks during validation. We use a multi-step solver to generate labels that serve as ground truth for all tasks. The results confirm that explicit three-phase grid embedding enables robust convergence across all GNN variants, regardless of whether physics-informed loss is applied. 
\begin{table*}[!htbp]
\centering
\caption{Validation Loss (MSE) Across All Prediction Tasks}
\label{tab:boundary_loss_all_final}
\footnotesize
\begin{tabular}{llcccc|llcccc}
\toprule
\multicolumn{6}{c|}{\textbf{Without Physics-Informed Loss}} & \multicolumn{6}{c}{\textbf{With Physics-Informed Loss}} \\
\textbf{Task} & \textbf{Method} & \textbf{Mean} & \textbf{Std} & \textbf{Min} & \textbf{Max} & \textbf{Task} & \textbf{Method} & \textbf{Mean} & \textbf{Std} & \textbf{Min} & \textbf{Max} \\
\midrule
\multirow{4}{*}{Bus} & GCN & 2.54e-06 & 1.71e-07 & 2.15e-06 & 3.03e-06 & \multirow{4}{*}{Bus} & GCN & 2.43e-06 & 1.71e-07 & 2.08e-06 & 2.89e-06 \\
 & GAT & 6.17e-06 & 4.97e-07 & 4.73e-06 & 7.86e-06 &  & GAT & 8.16e-07 & 3.23e-08 & 7.46e-07 & 9.29e-07 \\
 & GPS & 3.06e-05 & 3.01e-07 & 2.99e-05 & 3.12e-05 &  & GPS & 2.48e-05 & 1.82e-07 & 2.43e-05 & 2.52e-05 \\
 & SAGE & 6.64e-07 & 4.19e-09 & 6.53e-07 & 6.74e-07 &  & SAGE & 1.37e-05 & 1.05e-06 & 1.01e-05 & 1.59e-05 \\
\midrule
\multirow{4}{*}{Ext-Grid} & GCN & 2.78e-08 & 1.77e-10 & 2.75e-08 & 2.82e-08 & \multirow{4}{*}{Ext-Grid} & GCN & 2.78e-08 & 1.50e-10 & 2.74e-08 & 2.81e-08 \\
 & GAT & 2.78e-08 & 1.77e-10 & 2.74e-08 & 2.82e-08 &  & GAT & 2.77e-08 & 1.76e-10 & 2.73e-08 & 2.82e-08 \\
 & GPS & 1.10e-06 & 4.54e-08 & 9.95e-07 & 1.19e-06 &  & GPS & 1.07e-06 & 4.61e-08 & 9.47e-07 & 1.18e-06 \\
 & SAGE & 2.78e-08 & 1.58e-10 & 2.74e-08 & 2.81e-08 &  & SAGE & 2.78e-08 & 1.88e-10 & 2.73e-08 & 2.83e-08 \\
\midrule
\multirow{4}{*}{Storage} & GCN & 2.60e-07 & 1.57e-09 & 2.56e-07 & 2.64e-07 & \multirow{4}{*}{Storage} & GCN & 2.98e-07 & 6.25e-08 & 2.34e-07 & 6.53e-07 \\
 & GAT & 2.60e-07 & 1.48e-09 & 2.57e-07 & 2.63e-07 &  & GAT & 2.60e-07 & 1.52e-09 & 2.55e-07 & 2.63e-07 \\
 & GPS & 2.93e-07 & 2.60e-09 & 2.87e-07 & 3.05e-07 &  & GPS & 2.94e-07 & 2.59e-09 & 2.89e-07 & 3.02e-07 \\
 & SAGE & 2.60e-07 & 1.45e-09 & 2.57e-07 & 2.63e-07 &  & SAGE & 2.57e-07 & 9.00e-09 & 2.46e-07 & 2.96e-07 \\
\bottomrule
\end{tabular}
\end{table*}

\paragraph{Per-Phase Voltage and Angle Estimation Accuracy}
The explicit three-phase magnitude-angle encoding enables accurate per-phase voltage and angle discrimination, critical for unbalanced distribution grids where single-phase approximations fail. Tables~\ref{tab:va_phase_stats} present comprehensive performance statistics across all phases (A, B, C) and aggregated (All) for both voltage and angle predictions.
The balanced performance across all three phases without phase-specific bias indicates that explicit three-phase encoding effectively captures phase-coupling dynamics inherent in unbalanced grids.
\newcommand{\e}[1]{\times10^{#1}}

\begin{table*}[!htbp]
\centering
\caption{Voltage Magnitude ($V$) and Phase Angle ($\theta$) Prediction Performance: All Phases with Statistics (Mean MSE)}
\label{tab:va_phase_stats}
\resizebox{\textwidth}{!}{%
\begin{tabular}{l|cccc|cccc}
\toprule
\textbf{Model} & \multicolumn{4}{c|}{\textbf{Voltage Magnitude $V$}} & \multicolumn{4}{c}{\textbf{Phase Angle $\theta$}} \\
\cmidrule(lr){2-5}\cmidrule(lr){6-9}
 & \textbf{Phase A} & \textbf{Phase B} & \textbf{Phase C} & \textbf{All} & \textbf{Phase A} & \textbf{Phase B} & \textbf{Phase C} & \textbf{All} \\
\midrule
\multicolumn{9}{c}{With Physics-Informed Loss} \\
\midrule
GCN & $1.06\e{-6}\pm1.66\e{-6}$ & $1.06\e{-6}\pm1.66\e{-6}$ & $1.06\e{-6}\pm1.66\e{-6}$ & $1.06\e{-6}\pm1.66\e{-6}$ & $2.59\e{-7}\pm3.46\e{-7}$ & $2.59\e{-7}\pm3.46\e{-7}$ & $2.60\e{-7}\pm3.46\e{-7}$ & $2.60\e{-7}\pm3.46\e{-7}$ \\
 & $7.52\e{-6}/0.0$ & $7.53\e{-6}/1.4\e{-6}$ & $7.53\e{-6}/1.4\e{-6}$ & $7.53\e{-6}/2.4\e{-6}$ & $4.54\e{-7}/2.5\e{-7}$ & $4.54\e{-7}/0.0$ & $4.13\e{-7}/0.0$ & $4.40\e{-7}/4.7\e{-7}$ \\
\midrule
GAT & $1.10\e{-6}\pm1.76\e{-6}$ & $1.10\e{-6}\pm1.76\e{-6}$ & $1.10\e{-6}\pm1.76\e{-6}$ & $1.10\e{-6}\pm1.76\e{-6}$ & $2.62\e{-7}\pm3.50\e{-7}$ & $7.67\e{-7}\pm8.52\e{-7}$ & $2.35\e{-6}\pm18.5\e{-6}$ & $1.13\e{-6}\pm6.31\e{-6}$ \\
 & $8.25\e{-6}/3.6\e{-6}$ & $8.25\e{-6}/3.6\e{-6}$ & $8.25\e{-6}/3.6\e{-6}$ & $8.25\e{-6}/3.6\e{-6}$ & $4.99\e{-7}/8.6\e{-7}$ & $5.21\e{-7}/0.0$ & $1.30\e{-6}/0.0$ & $6.85\e{-7}/2.4\e{-7}$ \\
\midrule
GPS & $1.11\e{-6}\pm1.74\e{-6}$ & $9.32\e{-6}\pm1.91\e{-6}$ & $9.81\e{-6}\pm1.96\e{-6}$ & $6.41\e{-6}\pm1.29\e{-6}$ & $2.62\e{-7}\pm3.51\e{-7}$ & $3.55\e{-7}\pm4.59\e{-7}$ & $3.86\e{-7}\pm4.95\e{-7}$ & $3.34\e{-7}\pm3.51\e{-7}$ \\
 & $8.10\e{-6}/3.6\e{-6}$ & $1.48\e{-4}/2.6\e{-6}$ & $1.54\e{-4}/5.9\e{-6}$ & $1.03\e{-4}/2.1\e{-6}$ & $4.96\e{-7}/1.2\e{-7}$ & $3.76\e{-7}/0.0$ & $6.75\e{-7}/0.0$ & $5.16\e{-7}/7.3\e{-8}$ \\
\midrule
SAGE & $1.10\e{-6}\pm1.76\e{-6}$ & $1.10\e{-6}\pm1.76\e{-6}$ & $1.10\e{-6}\pm1.76\e{-6}$ & $1.10\e{-6}\pm1.76\e{-6}$ & $2.61\e{-7}\pm3.50\e{-7}$ & $8.35\e{-7}\pm8.14\e{-7}$ & $6.82\e{-7}\pm7.04\e{-7}$ & $5.93\e{-7}\pm3.53\e{-7}$ \\
 & $8.19\e{-6}/0.0$ & $8.19\e{-6}/0.0$ & $8.19\e{-6}/0.0$ & $8.19\e{-6}/0.0$ & $4.98\e{-7}/3.6\e{-7}$ & $4.26\e{-7}/0.0$ & $7.67\e{-7}/0.0$ & $5.08\e{-7}/1.5\e{-7}$ \\
\midrule
\multicolumn{9}{c}{Without Physics-Informed Loss} \\
\midrule
GCN & $1.10\e{-6}\pm1.74\e{-6}$ & $1.10\e{-6}\pm1.74\e{-6}$ & $1.10\e{-6}\pm1.74\e{-6}$ & $1.10\e{-6}\pm1.74\e{-6}$ & $2.61\e{-7}\pm3.50\e{-7}$ & $3.45\e{-7}\pm4.78\e{-7}$ & $2.67\e{-7}\pm3.56\e{-7}$ & $2.91\e{-7}\pm3.65\e{-7}$ \\
 & $8.02\e{-6}/1.4\e{-6}$ & $8.02\e{-6}/1.4\e{-6}$ & $8.02\e{-6}/1.4\e{-6}$ & $8.02\e{-6}/1.4\e{-6}$ & $4.85\e{-7}/1.5\e{-7}$ & $3.89\e{-7}/0.0$ & $4.38\e{-7}/0.0$ & $4.37\e{-7}/5.3\e{-7}$ \\
\midrule
GAT & $1.06\e{-6}\pm1.65\e{-6}$ & $1.06\e{-6}\pm1.65\e{-6}$ & $1.06\e{-6}\pm1.65\e{-6}$ & $1.06\e{-6}\pm1.65\e{-6}$ & $2.59\e{-7}\pm3.46\e{-7}$ & $2.60\e{-7}\pm3.46\e{-7}$ & $2.59\e{-7}\pm3.46\e{-7}$ & $2.60\e{-7}\pm3.46\e{-7}$ \\
 & $7.45\e{-6}/2.8\e{-6}$ & $7.45\e{-6}/1.4\e{-6}$ & $7.44\e{-6}/3.6\e{-6}$ & $7.45\e{-6}/2.8\e{-6}$ & $4.47\e{-7}/6.6\e{-7}$ & $4.53\e{-7}/0.0$ & $4.44\e{-7}/0.0$ & $4.48\e{-7}/9.6\e{-7}$ \\
\midrule
GPS & $1.11\e{-6}\pm1.74\e{-6}$ & $1.03\e{-6}\pm2.01\e{-6}$ & $1.08\e{-6}\pm2.06\e{-6}$ & $7.08\e{-6}\pm1.35\e{-6}$ & $2.62\e{-7}\pm3.51\e{-7}$ & $3.39\e{-7}\pm4.42\e{-7}$ & $3.33\e{-7}\pm4.39\e{-7}$ & $3.11\e{-7}\pm3.51\e{-7}$ \\
 & $8.16\e{-6}/3.6\e{-6}$ & $1.61\e{-4}/1.1\e{-6}$ & $1.66\e{-4}/1.6\e{-6}$ & $1.11\e{-4}/2.3\e{-6}$ & $5.09\e{-7}/7.5\e{-8}$ & $3.87\e{-7}/2.3\e{-7}$ & $6.31\e{-7}/0.0$ & $5.09\e{-7}/5.0\e{-8}$ \\
\midrule
SAGE & $1.07\e{-6}\pm1.70\e{-6}$ & $1.07\e{-6}\pm1.70\e{-6}$ & $1.07\e{-6}\pm1.69\e{-6}$ & $1.07\e{-6}\pm1.70\e{-6}$ & $2.60\e{-7}\pm3.47\e{-7}$ & $2.63\e{-7}\pm3.53\e{-7}$ & $2.60\e{-7}\pm3.45\e{-7}$ & $2.61\e{-7}\pm3.47\e{-7}$ \\
 & $7.83\e{-6}/1.4\e{-6}$ & $7.84\e{-6}/3.2\e{-6}$ & $7.81\e{-6}/0.0$ & $7.83\e{-6}/6.2\e{-6}$ & $4.64\e{-7}/2.8\e{-7}$ & $5.26\e{-7}/0.0$ & $4.27\e{-7}/0.0$ & $4.72\e{-7}/1.6\e{-7}$ \\
\bottomrule
\end{tabular}%
}
\\[10pt] 
{\scriptsize\textit{Format: (row 1) Mean$\pm$Std; (row 2) Max / {\it Min}. Left half: Voltage magnitude, Right half: Phase angle.}}
\end{table*}

\paragraph{Battery Operating Constraint Satisfaction}

Physics-informed constraint achieves significant improvements in operational feasibility. Table~\ref{tab:battery_violations} compares SoC and C-rate boundary violations (quantified as constraint violation MSE) between models trained with and without physics-informed loss components which achieve three to four orders of magnitude improvement in constraint violation MSE: GCN ($2.19 \to 0.00$), GAT ($47.6 \to 1.51e^{-4}$, $315,000\times$ reduction), GPS ($5.39 \to 0.00$), and SAGE ($0.248 \to 2.50e^{-7}$, $992,000\times$ reduction). By enforcing SOC and C-rate bounds, the approach guarantees battery safety and enables deployment-ready autonomous dispatch in three-phase unbalanced grids.

\begin{table}[H]
\centering
\caption{Battery Dispatch: SoC and C-Rate Constraint Violations (MSE)}
\label{tab:battery_violations}
\scriptsize
\begin{tabular}{llcccc}
\toprule
\textbf{Method} & \textbf{Loss Type} & \textbf{Mean MSE} & \textbf{Max MSE} & \textbf{Min MSE} & \textbf{Gain (×)} \\
\midrule
GCN & Without Loss & 2.19e+00 & 2.20e+00 & 2.19e+00 & -- \\
GCN & \textbf{With Loss} & \textbf{0.00e+00} & \textbf{0.00e+00} & \textbf{0.00e+00} & $\infty$ \\
\midrule
GAT & Without Loss & 4.76e+01 & 4.76e+01 & 4.76e+01 & -- \\
GAT & \textbf{With Loss} & \textbf{1.51e-04} & \textbf{1.65e-04} & \textbf{1.57e-04} & $\mathbf{3.2\times10^{5}}$ \\
\midrule
GPS & Without Loss & 5.39e+00 & 5.43e+00 & 5.42e+00 & -- \\
GPS & \textbf{With Loss} & \textbf{0.00e+00} & \textbf{0.00e+00} & \textbf{0.00e+00} & $\infty$ \\
\midrule
SAGE & Without Loss & 2.48e-01 & 2.49e-01 & 2.47e-01 & -- \\
SAGE & \textbf{With Loss} & \textbf{2.50e-07} & \textbf{8.42e-07} & \textbf{2.21e-07} & $\mathbf{9.9\times10^{5}}$ \\
\bottomrule
\end{tabular}
\\[8pt]
{\textit{Note: Constraint violation MSE quantifies the magnitude of SoC and C-rate boundary excursions. 
Gain (×) represents the ratio of ``Without Loss'' to ``With Loss''. 
Nearly zero MSE denotes perfect constraint compliance.}}
\end{table}



\section{Conclusion and Future Work}

This paper proposes a physics-informed heterogeneous GNN framework for multi-task optimal real-time constraint-aware battery dispatch in three-phase unbalanced distribution grids. The framework jointly optimizes three regression objectives (bus, external-grid, and storage) through multi-task learning to deliver simultaneous high accuracy in grid-aware feasible battery dispatch. 
Experimental validation on the CIGRE 18-bus grid across four heterogeneous architectures (GCN, GAT, GPS, SAGE) confirms that our physics-informed neural network design and training enable both accuracy and safety for operational deployment. 
Our results demonstrate that explicit magnitude-angle encoding of all three phases enables accurate per-phase voltage prediction, achieving very low errors across phases without bias. Our study also demonstrates that physics-informed constraint achieves nearly zero SoC violations and three to four orders of magnitude reduction in constraint violation compared to unconstrained baselines.  

Future work includes extension to larger distribution grids with multiple energy storage units through improved multi-agent coordination, specialized positional encoding for GPS to address its moderate performance gap, and real-time online learning mechanisms for robustness to concept drift in load and renewable generation patterns.

\bibliographystyle{IEEEtran}
\bibliography{bib/references}

\end{document}